\documentclass{article}

\PassOptionsToPackage{numbers, compress}{natbib}
\usepackage[final,nonatbib]{tackling_climate_workshop_style}

\usepackage[utf8]{inputenc} % allow utf-8 input
\usepackage[T1]{fontenc}    % use 8-bit T1 fonts
\usepackage{hyperref}       % hyperlinks
\usepackage{url}            % simple URL typesetting
\usepackage{booktabs}       % professional-quality tables
\usepackage{amsfonts}       % blackboard math symbols
\usepackage{nicefrac}       % compact symbols for 1/2, etc.
\usepackage{microtype}      % microtypography

\usepackage{graphicx}
\usepackage{natbib} % Use NatBib
\title{Predicting Atlantic Multidecadal Variability}

\author{
    Glenn Liu \\
    EAPS, MIT\\
    Department of Physical Oceanography, WHOI\\
    \And
    Peidong Wang \\
    EAPS, MIT\\
    \And
    Matthew Beveridge \\
    EECS, MIT\\
    \And
    Young-Oh Kwon \\
    Department of Physical Oceanography, WHOI\\
    \And
    Iddo Drori\\
    EECS, MIT\\
}

\begin{document}

\maketitle

\begin{abstract}
Atlantic Multidecadal Variability (AMV) describes variations of North Atlantic sea surface temperature with a typical cycle of between 60 and 70 years. AMV strongly impacts local climate over North America and Europe, therefore prediction of AMV, especially the extreme values, is of great societal utility for understanding and responding to regional climate change. This work tests multiple machine learning models to improve the state of AMV prediction from maps of sea surface temperature, salinity, and sea level pressure in the North Atlantic region. We use data from the Community Earth System Model 1 Large Ensemble Project, a state-of-the-art climate model with 3,440 years of data. Our results demonstrate that all of the models we use outperform the traditional persistence forecast baseline. Predicting the AMV is important for identifying future extreme temperatures and precipitation, as well as hurricane activity, in Europe and North America up to 25 years in advance.
\end{abstract}

\section{Introduction}
\label{sec:introduction}
The Atlantic Multidecadal Variability (AMV, also known as the Atlantic Multidecadal Oscillation) is a basin-wide fluctuation of sea-surface temperatures (SST) in the North Atlantic with a periodicity of approximately 60 to 70 years. AMV has broad societal impacts. The positive phase of AMV, for example, has been shown to be associated with anomalously warm summers in northern Europe and hot, dry summers in southern Europe\cite{Gao_2019}, and increased hurricane activity \cite{Zhang_Delworth_2006}. These impacts highlight the importance of predicting extreme AMV states. 

\begin{figure}
    \centering
    \includegraphics[width=4.5in]{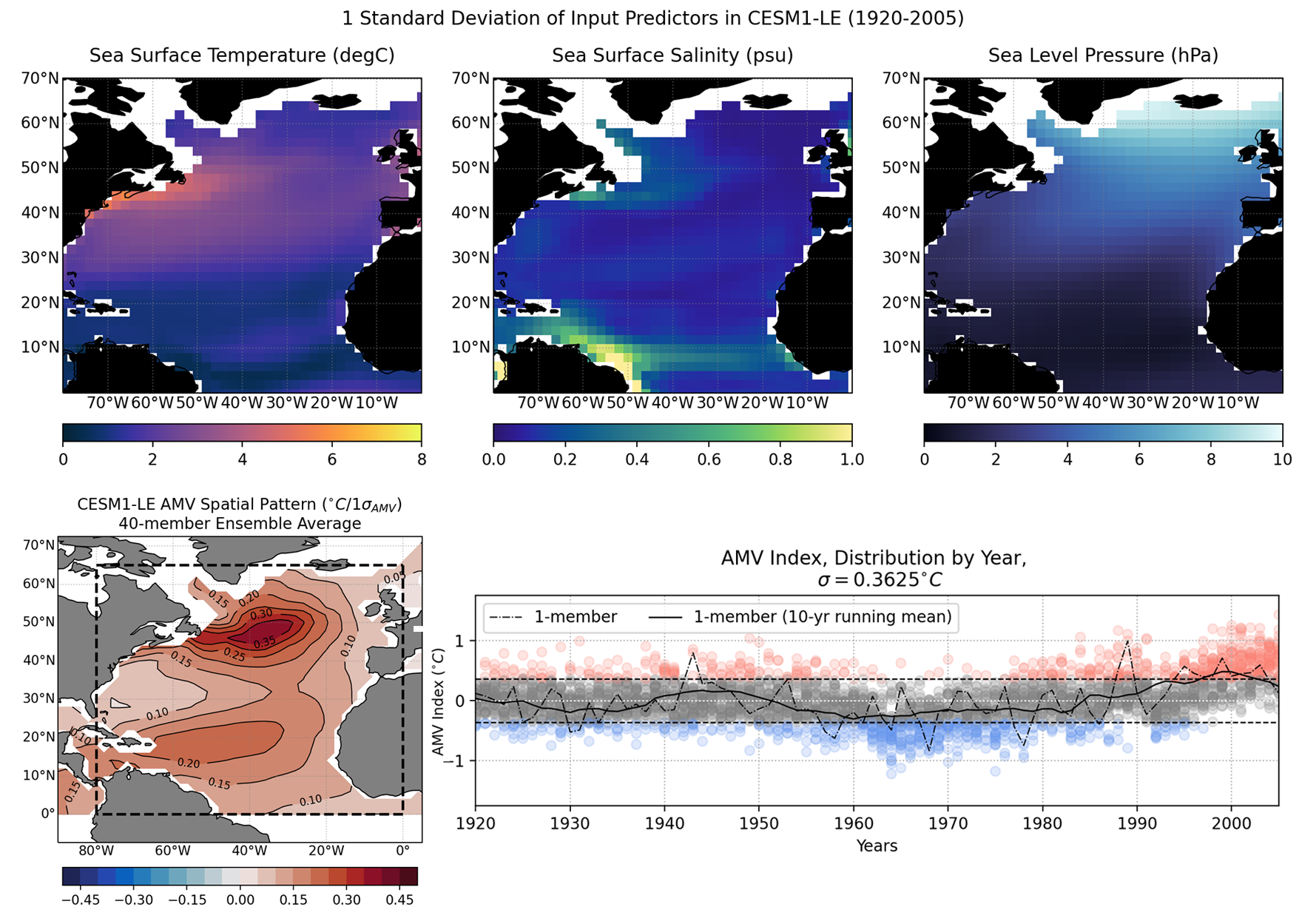}
    \caption{Variability of input predictors, which include sea surface temperature, sea surface salinity, and sea level pressure. The prediction objectives (lower right) are strong positive (red) and negative (blue) AMV states outside 1 standard deviation of the AMV Index (dashed black line). The AMV spatial pattern from the CESM simulation reasonably captures the enhanced warming at subpolar and tropical latitudes.}
    \label{fig:cesm_data}
\end{figure}

The state of AMV is measured by the AMV Index (Figure \ref{fig:cesm_data} bottom right panel, black solid line), calculated by averaging sea-surface temperature (SST) anomalies over the entire North Atlantic basin. The spatial pattern of SST associated with a positive AMV phase is characterized with the maximum warming in the subpolar North Atlantic and a secondary maxima in the tropical Atlantic with minimum warming (or slightly cooling) in between.

Notwithstanding the value of reliable prediction of AMV, progress in predicting AMV at decadal and longer time scales has been limited. Previous efforts have used the computationally expensive numerical climate models to perform seasonal to multi-year prediction with lead times of up to 10 years. The subpolar region in the North Atlantic has been shown to be one of the most predictable regions over the globe, and associated with the predictability of weather and climate in Europe and North America up to 10 years. An outstanding question is whether such predictability can be extended to prediction lead times longer than 10 years, particularly in a changing climate. 

Our objective is to predict these extreme states of the AMV using various oceanic and atmospheric fields as predictors. This is formulated as a classification problem, where years above and below 1 standard deviation of the AMV Index correspond to extreme warm and cold states. In this work, we use multiple ML models to explore the predictability of AMV up to 25 years in advance.

\paragraph{Related Work.} ML techniques have been successfully applied to predict climate variability, especially the El Niño-Southern Oscillation (ENSO). ENSO is an interannual mode of variability (each cycle is about 3-7 years) in the tropical Pacific Ocean. Several studies have used CNN to predict ENSO 12 to 16 months ahead using various features (e.g. SST, ocean heat content, sea surface height etc. \cite{Ham_2020,Pal_2020,Yan_2020}. This outperformed the typical 10-month lead time ENSO forecast with state-of-the-art, fully-coupled dynamical models \cite{Ham_2020}.

However, little work has been done to predict decadal and longer modes of variability such as the AMV using ML. The biggest challenge is the lack of data. Widespread observational records for many variables are only available after the 1980s, limiting both the temporal extent and pool of predictors that may be used for training. For interannual modes such as ENSO, current observations can be easily partitioned into ample training and testing datasets with multiple ENSO cycles in each subset of data. However, a single AMV cycle requires 60-70 years, making it nearly impossible to train and test a neural network on observational data alone.

\section{Data and Methods}

To remedy the lack of observational data for AMV, we used the Community Earth System Model version 1.1 Large Ensemble Simulations\footnote{\href{https://ncar.github.io/cesm-lens-aws/}{Available to download at https://ncar.github.io/cesm-lens-aws/}}.
This is a fully-coupled global climate model that includes the best of current knowledge of physics, and has shown good performance in simulating the periodicity and large-scale patterns of AMV comparable with observations \cite{Wang_2015}. There are a total of 40 ensemble runs, each between 1920 and 2005. The individual runs are slightly perturbed in their initial conditions and thus treated as 40 parallel worlds. The variability of ocean and atmospheric dynamics in each run represents intrinsic natural variability in the climate system that we aim to predict, and provides a diverse subsampling of AMV behavior. 

Our objective is to train machine learning models to predict the AMV state (AMV+, AMV-, Neutral). Each model is given 2-dimensional maps of SST, sea surface salinity (SSS), and sea-level pressure (SLP), and is trained to predict the AMV state at a given lead time ahead, from 0-year (AMV at the current year) to 25-year lead (AMV 25 years into the future). We train models to make predictions every 3-years. This results in 9 models for each architecture, each specialized in predicting AMV at a particular lead time. The procedure is repeated 10 times for each lead time to account for sensitivity to the initialization of model weights and randomness during the training and testing process.

To quantify the success of each model, we define prediction skill as the accuracy of correct predictions for each AMV state. We compare the performance of the models against a persistence forecast, which is a traditional baseline in the discipline. The persistence forecast is formulated such that the current state is used as a prediction for the target state. The accuracy for this prediction method is evaluated for each lead time in the dataset.

We used a convolutional neural network (CNN), residual neural network (specifically ResNet50), AutoML and FractalDB in this study. A detailed description of the hyperparameters associated with each ML model and the pre-processing of the data are included in the appendix. 

\section{Results}

\begin{figure}
    \centering
    \includegraphics[width=5in,scale=0.90]{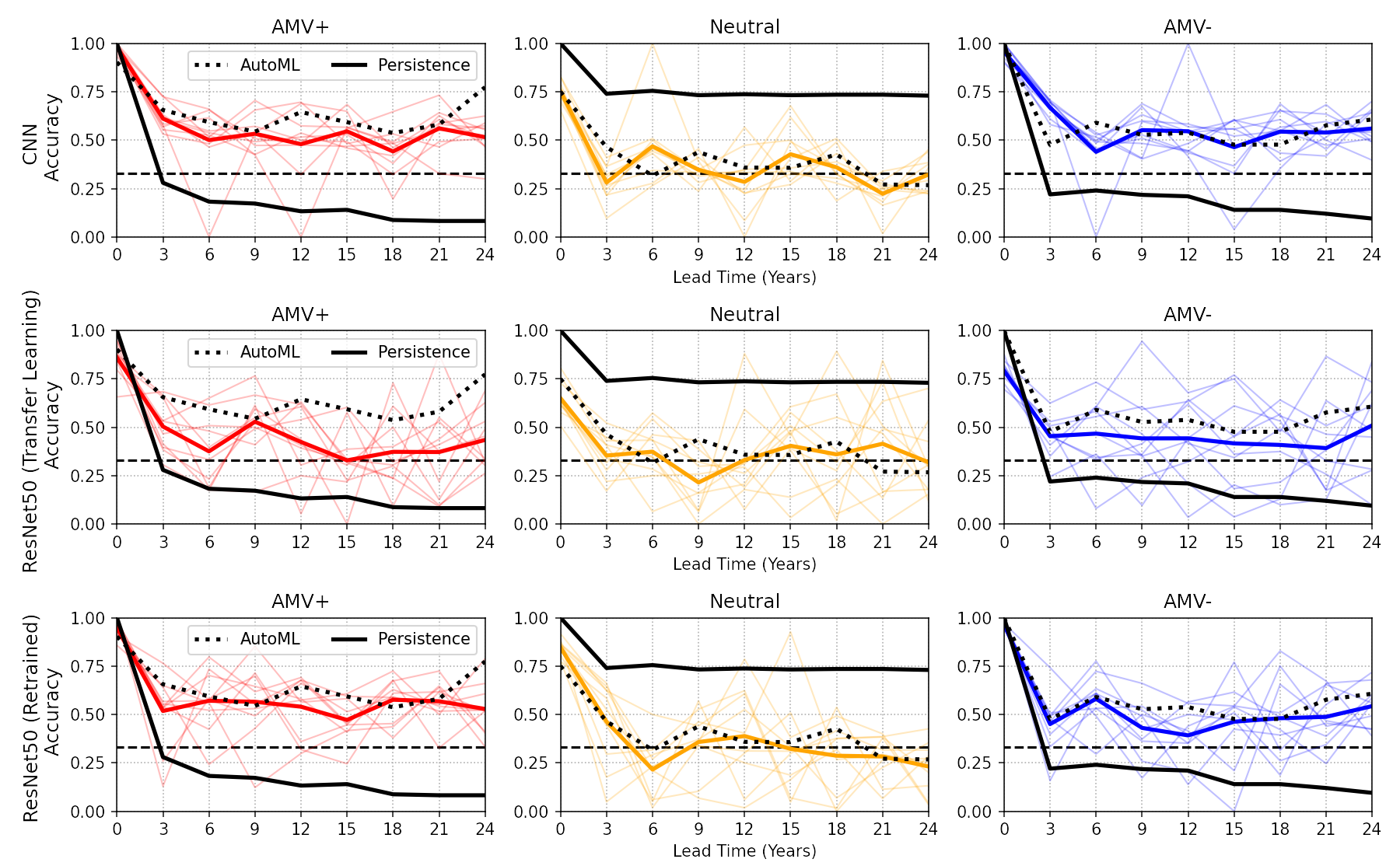}
    \caption{Accuracy of AMV state predictions by lead year for each class (columns) and each machine learning model (rows). AutoML results (dotted black line), persistence baseline (solid black line), and random chance (33\% for each class, dashed black line) are shown on each subplot for comparison. For each of the CNN and ResNet50, we performed 10 ensemble runs with different initial conditions, each of the ensemble is shown in the thin colored lines, and the thick colored lines indicate ensemble mean.}
    \label{fig:results_byclass}
\end{figure}

\paragraph{Prediction Skill by AMV States.} Fig. \ref{fig:results_byclass} shows the accuracy of AMV prediction for positive, negative, and neutral states from CNN, ResNet50 (transfer learning, fully retrained) and AutoML, compared with the persistence baseline and random guess baseline. Our results show that extreme AMV states (both positive and negative phases) are predictable well beyond a 10-year lead time, further than the limit explored by previous research. All ML models outperform the persistence baseline and the accuracy expected due to random chance. The difference in skill is more pronounced at longer lead times. It is more important to achieve higher predictive skill for the extreme AMV phases as they may lead to more pronounced societal impacts. The models perform poorly for neutral AMV states, and are well below the persistence baseline. This arises from the challenge of predicting the neutral AMV state. It is more common for the AMV Index to maintain neutral conditions at any given year rather than to transition into an extreme warm or cold state, contributing to the high skill seen in the persistence forecasts.

One surprising aspect is that the models are able to skillfully predict both positive and negative AMV states. While higher predictability for positive AMV states can be explained the presence of the global warming trend in SST, comparable skill is achieved for negative AMV states. This suggests that the negative AMV states must have an equivalent signature in the predictors that contribute to this increased skill. The nature and origin of this signature for negative AMV states is a subject which requires further investigation.

\paragraph{Differences in Skill by Model.} Overall, AutoML demonstrates higher predictive skill compared to the persistence baseline and performs better than the CNN and ResNet50 baselines. This difference is more pronounced for predicting positive AMV states. AutoML outperforms the CNN and ResNet50 by up to 25\% at a prediction lead time of 24 years. An improvement in prediction skill across all lead times is achieved by fully-retraining all weights of the ResNet50 compared to the traditional transfer learning approach. As with previous results, the differences between these two approaches become more pronounced at long prediction lead times, and for the prediction of positive AMV states. Similar accuracies are achieved by both the fully retrained ResNet50 and CNN, with around a 50\% accuracy for a prediction of AMV state with a 25 year lead time. AutoML, with minimal user-end tuning, is able to outperform both the ResNet50 and CNN --- both of which required testing and selection of hyperparameters. The ability of this method to work with minimal user involvement highlights its adaptability to domains such as climate science and prediction and potential for application to other challenges and problems in climate science. In the spirit of reproducible research we make our code and data publicly available online\footnote{\href{https://www.dropbox.com/s/d2wokw4f15mejlk/CESM_data.zip?dl=0}{https://www.dropbox.com/s/d2wokw4f15mejlk/CESM\_data.zip?dl=0}; \href{https://www.dropbox.com/s/uuihbwwovr160fm/predict_amv.zip?dl=0}{ https://www.dropbox.com/s/uuihbwwovr160fm/predict\_amv.zip?dl=0}}.

\section{Conclusions}

In summary, our results highlight the potential for applying ML to the problem of decadal climate predictability. Specifically, we found that our ML models are able to predict AMV extreme states up to 25 years into the future. The models outperform the traditional persistence baseline, especially at long prediction lead times. This suggests that the features identified by the ML models from the oceanic and atmospheric data provide potential for decadal-timescale forecast ability. Furthermore, we found that climate data present a new challenge for ML models. Retraining all weights for the pre-trained ResNet50 model yielded improved accuracy compared to using a transfer learning approach, suggesting that the features learned from traditional ML databases such as ImageNet do not necessarily translate to skill in climate prediction. Further investigation is needed to interpret the sources of AMV predictability learned by these methods. For example, employing methods such as Class Activation Mapping \cite{zhou2015cnnlocalization} to identify which region contributes most to the prediction is one avenue for future work. This would reveal if particular spatial patterns in the predictor variables are critical for skillful prediction of extreme AMV states. Additionally, examining these patterns and their correspondence with the known processes that drive AMV may potentially allow for the discovery of unknown linkages and interactions between these processes, further boosting our understanding of long-term climate predictability and change using ML techniques.

As the AMV has been shown to be associated with various weather and climate phenomena around the North Atlantic, the prediction of AMV extreme states, out to 25 years, has great socioeconomic value. In particular, state-of-the-art predictions using the current generation climate models only extend 10-years out into the future, and beyond that the general practice is only to consider the climate projection in response to the climate change scenarios near the end of the 21\textsuperscript{st} century. For the prediction, effects of both the natural variability and climate change should be considered together as shown by our result. Therefore, our finding that ML models can produce a skillful prediction with lead time beyond 10 years out to 25 years extends the prediction horizon to an unprecedented stage. Our results suggest the potential of applying ML to other challenging problems of long-term prediction in the climate system, from similar decadal climate modes such as the Indian Ocean Dipole or the variability in sea ice coverage over the polar regions.

\bibliography{bibliography}

\newpage
\clearpage

\section*{Appendix}
\begin{figure}
    \renewcommand{\thefigure}{A1}
    \centering
    \includegraphics[width=0.6\textwidth]{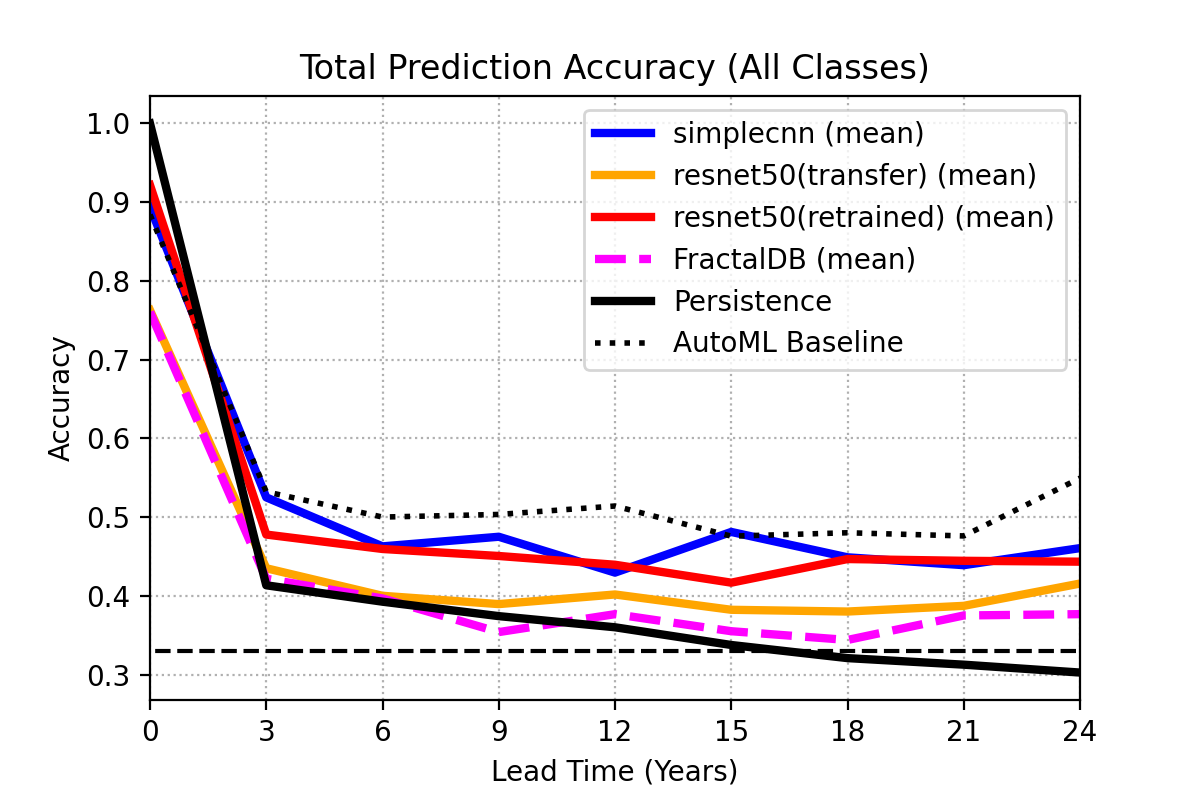}
    \caption{Comparison of results for different architectures: Mean prediction accuracy by lead year for each machine learning model. At long lead times ($\geq$ 18 years), AutoML outperforms both machine learning and persistence baselines.}
    \label{fig:results_bymodel}
\end{figure}

We provide additional implementation details, describe the architectures in detail, and show a comparison of results using the different architectures in Figure \ref{fig:results_bymodel}.

\paragraph{Feature Selection and Pre-processing.} Previous studies have suggested various drivers of AMV. One viewpoint is that the AMV is primarily driven by ocean dynamics, especially by changes in the Atlantic Meridional Overturning Circulation --- the oceanic pathway of equator-to-pole heat transport \cite{Zhang_2019}. Another viewpoint is that AMV is driven by atmospheric dynamics, particularly by the North Atlantic Oscillation \cite{Clement_2015}. The North Atlantic Oscillation is characterized by variability in sea level pressure that modulates the strength of jet stream, therefore affecting the heat exchange between the atmosphere and ocean. Considering the contribution of each of these drivers is important for skillful prediction of the AMV state; we will use both oceanic and atmospheric drivers in the AMV prediction in this study. 

The input features are yearly snapshots of sea surface temperature (SST), sea surface salinity (SSS), and sea level pressure (SLP) (Figure \ref{fig:cesm_data}, top row). Previous studies have shown that SSS varies coherently with AMV on multi-decadal timescales; since SSS variability is controlled primarily through oceanic processes, this variable provides an opportunity to evaluate how ocean drivers contribute to AMV predictability \cite{Zhang_2017_B}. SLP anomalies are directly involved in calculating the state of North Atlantic Oscillation --- the leading mode of atmospheric variability in the Atlantic sector and one of the theorized AMV drivers --- and thus provide insight for atmospheric drivers of AMV. 

All inputs are normalized and standardized prior to input, and re-gridded to 224 x 224 resolution using the nearest-neighbor method to prepare for input into our models. We also apply the land and sea ice masks from ocean variables (SST, SSS) to the atmospheric variable (SLP) so the extra information from land in SLP does not bias the prediction. Note that the long-term trend is not removed from any of the variables, as we intend to forecast the signal due to both the climate change and natural variability, which is appropriate considering long lead times considered here.

\paragraph{AMV Classification.} Extreme AMV states often have a disproportionate impact on the climate system. To this end, we formulate AMV prediction as a classification problem to focus on predicting these extreme positive states and negative states, instead of trying to predict the exact AMV indices as a regression problem. The AMV Index is first calculated by taking the area-weighted average of SST anomalies between 80$^{\circ}$W to 0$^{\circ}$ Longitude and 0$^{\circ}$ to 65$^{\circ}$N Latitude (Figure \ref{fig:cesm_data}). Each year is then classified as an extreme positive (AMV+), negative (AMV-), or neutral AMV state if the AMV Index is above, below, or within 1 standard deviation of the AMV Index (0.3625$^{\circ}$C). Due to the large number of neutral states, an equal number of 300 samples are randomly selected from each class and shuffled prior to input to maintain a similar distribution among all three categories. We use 80\% of the samples for the training set, and the remaining 20\% as the validation and testing set.

\paragraph{Convolutional neural network.}
We use a 2-layer Convolutional Neural Network (CNN) to predict the AMV index. Each input variable is treated as a separate channel, and each feature consists of the dimensions \textit{channel} (3) x \textit{latitude} (224) x \textit{longitude} (224) (equivalent of having a horizontal resolution of 0.25\textsuperscript{o} $\times$ 0.33\textsuperscript{o}). The first convolutional filter (of size of 2x3) and max pooling (of size 2x3) outputs 32 feature maps. This rectangular filter size was selected to capture zonal (East-West) temperature patterns associated with the AMV. A second convolutional layer (of size 3x3) and max pooling layer (of size 2x3) receives the 32 feature maps as input and outputs 64 feature maps. In each convolutional layer, we apply a ReLU activation after convolution. We then flatten the data and pass it through 2 fully-connected layers before making a final prediction of the AMV index. Each fully-connected layer is also followed by a ReLU activation. We train the CNN for 20 epochs, using early stopping, halting the training process when validation loss increases for 3-consecutive epochs. We experimented with different number of CNN layers and number of neurons in each layer. The CNN tends to overfit on the training set as we increase the number of layers and neurons. We used the same CNN architecture but for the re-gridded data (to 244 by 244 pixels); the results do not change when using the input data at its original resolution (33 by 41). 

\paragraph{Residual neural network.} We also train a model based on a pre-trained ResNet50 \citep{resnet50}. Since ImageNet models have not been previously used on climate model data for AMV prediction at long timescales, it is necessary to examine the suitability of transfer learning. We thus investigate two different approaches to training ResNet50. The first is the traditional transfer learning scenario, where only the weights of the last layer are unfrozen. The second approach involves unfreezing and retraining all weights within the network. In both cases, the last fully connected layer is adjusted to output 3 values instead of 1000 to align with our AMV classification objective. Each model is trained for 20 epochs with early stopping for 3-epochs of consecutive validation loss increases.

\paragraph{FractalDB.} Recent work has shown that CNNs pre-trained on a database of synthetic fractals (FractalDB) without natural images partially outperformed CNNs pre-trained on ImageNet-1k and Places-365, suggesting that naturally-occurring fractal patterns are useful for classification and prediction tasks \citep{kataoka2020pre}. In the climate sciences, one application of fractal theory is through the use of fractal dimensional analysis of temperature and precipitation time series to identify changes in behavior of climate modes such as ENSO \citep{nunes2011fractal}. While this suggests connections between self-similarity and climate variability in the time domain, corresponding relationships in the space domain have not been extensively explored, particularly in the context of multidecadal predictability. To this end, we investigate if ResNet50 pre-trained with a database of 1,000 fractal categories (FractalDB) offers improvements in AMV prediction skill compared to the conventional ResNet50 trained on ImageNet \footnote{ \href{https://github.com/hirokatsukataoka16/FractalDB-Pretrained-ResNet-PyTorch}{Available to download at: https://github.com/hirokatsukataoka16/FractalDB-Pretrained-ResNet-PyTorch}}. The same transfer-learning procedure describe in the ResNet section above is applied to fine-tune the weights of the final layer for the FractalDB case.

Pre-training ResNet50 using FractalDB leads to slightly worse prediction skill compared with the traditional ResNet50 pre-trained with ImageNet. This suggests not using natural images in pre-training ResNet50 does not lead to a large difference in performance when using the transfer learning approach. A more substantial difference is observed when re-training all weights of the network.

\paragraph{Automated machine learning.}
We apply auto-sklearn \citep{autosklearn} to the data as described in the Feature Selection and Pre-processing section. We use AutoML for finding the best models or ensembles at each lead time, resulting in total of 9 optimal pipelines at all prediction lead times. A typical pipeline consists of feature extraction, feature selection, and classification primitives. Our results yield diverse pipelines including margin based estimators, as well as random forests. The search time for each lead time is one hour, this allows sufficient time for model fitting. Unlike the ResNet50 and CNN, the predictors are flattened prior to input.

\end{document}